\documentclass[conference]{IEEEtranSPMB}

\IEEEoverridecommandlockouts
\ifCLASSINFOpdf
\else
\fi
\usepackage{amsmath,graphicx,soul,subcaption,balance}
\usepackage{balance}
\usepackage[numbers,sort&compress]{natbib}

\usepackage{parskip}
\usepackage{float} 

\usepackage{mathptmx}

\usepackage{multirow} 

\usepackage{color}
\usepackage{soul,mathrsfs}
\usepackage[utf8]{inputenc}
\usepackage[T1]{fontenc}
\usepackage{array}
\usepackage[hyphens, spaces, obeyspaces]{url}

\usepackage{hyperref}
\hypersetup{
    colorlinks=true,
    linkcolor=blue,
    filecolor=magenta,      
    urlcolor=cyan,
}

\usepackage[papersize={8.5in,11in}, left=1in, right=1in, top=1in, bottom=1in]{geometry}
\newcolumntype{L}[1]{>{\raggedright\let\newline\\\arraybackslash\hspace{0pt}}m{#1}}
\newcolumntype{C}[1]{>{\centering\let\newline\\\arraybackslash\hspace{0pt}}m{#1}}
\newcolumntype{R}[1]{>{\raggedleft\let\newline\\\arraybackslash\hspace{0pt}}m{#1}}

\usepackage[singlelinecheck=false,justification=centering]{caption}
\DeclareCaptionLabelFormat{mylabel}{#1 #2.\hspace{0.7ex}}
\renewcommand{\thetable}{\arabic{table}}

\usepackage{xcolor,colortbl}
\definecolor{light-gray}{gray}{0.83}
\newcommand{\GREY}{\cellcolor{light-gray}\bf} 

\newcommand{\spmbtitlefont}{\large\bf\vspace{2em}}

\newcommand{\subparagraph}{}
\usepackage[compact]{titlesec}
\titleformat{\section}
   {\center\normalfont\sc}{\thesection.}{0.7ex}{}
\titlespacing{\section}{0pt}{2ex}{1.5ex}
\titlespacing{\subsection}{0pt}{1.5ex}{1.2ex}
\titlespacing{\subsubsection}{0pt}{1ex}{0.9ex}
\makeatletter
\renewcommand*{\@seccntformat}[1]{\csname the#1\endcsname .\hspace{0.7em}}
\makeatother

\usepackage{fancyhdr,lastpage}
\fancypagestyle{firststyle}
{
   \fancyhf{}
	\lfoot{IEEE SPMB 2020}\rfoot{v1.0: June 1, 2020}
}
\title{\spmbtitlefont Seizure Type Classification using EEG signals and Machine Learning: Setting a benchmark{\vspace{-2.3\baselineskip}}}

    \author{\IEEEauthorblockN{
    Subhrajit Roy\textsuperscript{\it 1}, 
    Umar Asif\textsuperscript{\it 2}, 
    Jianbin Tang\textsuperscript{\it 2} and
    Stefan Harrer\textsuperscript{\it 2}
    }
    \vspace{0.5em}
    \IEEEauthorblockA{
        1. IBM Research Australia, now with Google Health, London, UK\\
        2. IBM Research Australia, Melbourne, VIC, AU \\
        roy.subhrajit20@gmail.com,  \{umarasif, jbtang, sharrer\}@au1.ibm.com
    }
}

\newcommand{\AbstractSummary}{S.\ Roy, et al.: Seizure Type ...}

\begin{document}

\IEEEaftertitletext{}
\maketitle

\begin{abstract}

Accurate classification of seizure types plays a crucial role in the treatment and disease management of epileptic patients. Epileptic seizure types not only impact the choice of drugs but also the range of activities a patient can safely engage in. 
With recent advances being made towards artificial intelligence enabled automatic seizure detection, the next frontier is the automatic classification of seizure types. 	
On that note, in this paper, we explore the application of machine learning algorithms for multi-class seizure type classification. 
We used the recently released TUH EEG seizure corpus (V1.4.0 and V1.5.2) and conducted a thorough search space exploration to evaluate the performance of a combination of various pre-processing techniques, machine learning algorithms, and corresponding hyperparameters on this task. 
We show that our algorithms can reach a weighted $F1$ score of up to 0.901 for seizure-wise cross validation and 0.561 for patient-wise cross validation thereby setting a benchmark for scalp EEG based multi-class seizure type classification.
\end{abstract}

{\textbf{\textit{keywords: Seizure type classification, Machine learning, Electroencephalography}}}

\IEEEpeerreviewmaketitle    
\thispagestyle{firststyle}  
\section{Introduction}
\label{sec:intro}
Despite many new advances in drug therapy and disease understanding, our capabilities in treating and managing epilepsy are extremely limited. Roughly 1\% of the world’s population, 65 million people, suffer from epilepsy \cite{epibasics}. For one third of these patients, no medical treatment options exist. These patients need to find ways to live with their condition and manage their daily lives around it. For the remaining two thirds of the patient population, medical treatment options are available but have vastly differing and constantly changing results and quality of treatment. These shortcomings in diagnosis and treatment options are caused by the fact that epilepsy is a highly individualized condition, i.e. it does not look the same in all patients and even for an individual patient disease expression changes over time. As a result, until recently, the lack of data and measurements made the correct matching of patients and drugs into an unnecessary, long process of trial and error. Manual diaries are the basic data source, but these have been proven to be only 50\% accurate. 

With the advent of mobile devices that allow to collect patient information in real-time, continuously and at the point of sensing, and leveraging miniaturization and IoT data collection platforms, new efforts are being directed towards building individualized patient management systems. Data that is more accurate and more extensive can be used to gain a patient specific understanding of the disease and provide support for decision-making in managing it.

Machine Learning has been successfully used to address a large variety of problems in the biomedical field, ranging from image classification in cancer diagnosis to the automatic interpretation of electronic health records. Recently, we reported results demonstrating feasibility of using specialized neural networks to classify EEG data into normal/abnormal EEG \cite{roy2018deep} and to automatically detect and predict seizures \cite{kiral2018epileptic}. In this paper, we expand on this work and discuss the feasibility of using machine learning algorithms for automatically distinguishing between different types of seizures as they are detected. This technology could support automatic, patient-specific seizure type logging in digital seizure diaries. Such seizure diaries could then be used to improve the performance of clinical trials through more efficient and reliable patient monitoring for endpoint detection, adherence control and patient retention \cite{harrer2019TIPS}. 

\section{Datasets}

We used the TUH EEG Seizure Corpus (TUSZ) \cite{shah2018temple}, which is the largest open source corpus of its type. This dataset includes the time of occurrence and type of each seizure. 

The dataset covers a total of 8 different types of seizures:  
Focal Non-Specific Seizure (FNSZ): Focal seizures not further specified by type; 
Generalized Non-Specific Seizure (GNSZ): Generalized seizures not further classified into one of the groups below; 
Simple Partial Seizure (SPSZ): Partial seizures during consciousness; Type specified by clinical signs only;
Complex Partial Seizure (CPSZ): Partial Seizures during unconsciousness; Type specified by clinical signs only;
Absence Seizure (ABSZ): Absence Discharges observed on EEG; patient loses consciousness for few seconds (Petit Mal);
Tonic Seizure (TNSZ): Stiffening of body during seizure (EEG effects disappear);
Tonic Clonic Seizure (TCSZ) : At first stiffening and then jerking of body (Grand Mal) and
Myoclonic Seizure (MYSZ): Myoclonus jerks of limbs.

v1.4.0 of the dataset released in Oct 2018 contains 2012 seizures as shown in Table \ref{table:data_stats_v1.4.0}. v1.5.2 of the dataset released in May 2020 contains 3050 seizures as shown in Table \ref{table:data_stats_v1.5.2}. 
Since the number of MYSZ samples was too low for statistically meaningful analysis, we did not include MYSZ seizures in our study hence making it a 7-class classification problem. 

\begin{table}[b!]
\vspace{-2em} 
\centering
\caption{Seizure Type Statistics for v1.4.0 }
\label{table:data_stats_v1.4.0}
 \begin{tabular}{|L{0.435\linewidth}|C{0.115\linewidth}|R{0.13\linewidth}|R{0.115\linewidth}|} 
 \hline
{\GREY Seizure Type} & {\GREY Seizure Number} & {\GREY Duration (Seconds)}  & {\GREY Patient Number}\\\hline
Focal Non-Specific (FNSZ) & 992 & 73466 & 109	\\\hline
Generalized Non-Specific (GNSZ) & 415	& 34348 & 44	\\\hline
Complex Partial (CPSZ) & 342 & 33088	& 34	\\\hline
Absence (ABSZ) & 99	& 852 & 13	\\\hline
Tonic (TNSZ) & 67	& 1271	& 2    \\\hline
Tonic Clonic (TCSZ) & 50	& 5630	& 11	\\\hline
Simple Partial (SPSZ) & 44	& 1534	& 2	  \\\hline
Myoclonic (MYSZ) & 3	& 1312	& 2     \\\hline
\end{tabular}
\vspace{-1em} 
\end{table}

\vspace{-2em}
\begin{table}[b!]
%
\centering
\caption{Seizure Type Statistics for v1.5.2 }
\label{table:data_stats_v1.5.2}
 \begin{tabular}{|L{0.435\linewidth}|C{0.115\linewidth}|R{0.13\linewidth}|R{0.115\linewidth}|} 
 \hline
{\GREY Seizure Type} & {\GREY Seizure Number} & {\GREY Duration (Seconds)}  & {\GREY Patient Number}\\\hline
Focal Non-Specific (FNSZ) & 1836 & 121139 & 150	\\\hline
Generalized Non-Specific (GNSZ) & 583	& 59717 & 81	\\\hline
Complex Partial (CPSZ) & 367 & 36321	& 41	\\\hline
Absence (ABSZ) & 99	& 852 & 12	\\\hline
Tonic (TNSZ) & 62	& 1204	& 3    \\\hline
Tonic Clonic (TCSZ) & 48	& 5548	& 14	\\\hline
Simple Partial (SPSZ) & 52	& 2146	& 3	  \\\hline
Myoclonic (MYSZ) & 3	& 1312	& 2     \\\hline
\end{tabular}
\end{table}

\vspace{2em} 
\section{Methods}
\label{sec:ml-algos}
In this section, we briefly discuss the data preparation strategies, pre-processing techniques, machine learning algorithms and hyperparameter tuning methodologies we have explored.

For pre-processing the dataset, we used two-popular methods which have been reported to be effective in analysing EEG signals \cite{paul2018various, schindler2007assessing}. In Method 1, we applied Fast Fourier Transform (FFT) to each $W_l$ seconds of clip having $O$ seconds overlap across all EEG channels. Next, we took $log_{10}()$ of the magnitudes of frequencies in the range $1-f_{max}$ Hz. After this operation, the dimension of each training sample becomes $(N, 47)$ where $N$ is the number of EEG channels. For Method 2, first FFT is applied to each $W_l$ seconds of clip having $O$ seconds overlap across all EEG channels. Next, the output of FFT is then clipped from 1 to $f_{max}$ Hz and  normalized across frequency buckets. The correlation coefficients $(N, N$) matrix is calculated from this normalized matrix of $(N, 47)$. Real eigenvalues are calculated on this correlation coefficients matrix with complex eigenvalues made real by taking the complex magnitude.  We only considered the upper right triangle of the $(N, N)$ correlation coefficients matrix (since it is symmetric) and sorted the eigenvalues by magnitude.

For classification, we used the following algorithms: k-Nearest Neighbors (k-NN), Stochastic Gradient Descent (SGD),  XGBoost, and Convolutional Neural Networks (CNN). For the first three algorithms, we used HyperOpt \cite{bergstra2015hyperopt} to choose the best hyperparameters. For CNN models, we used the popular ResNet50 \cite{he2016deep} model and retrained the final layer for this task. 

For cross validation, in v1.4.0, TNSZ and SPSZ classes only contain data from 2 patients therefore, patient-wise cross validation will not yield statistically meaningful results. 
Hence previous work in the field \cite{ahmedt2019neural, asif2019seizurenet} chose to apply 5-fold seizure-wise cross validation, in which the seizures from different seizure types will be equally and randomly allocated to 5 folds. In this scenario train and test datasets can contain different seizure samples from the same patient. 
Since v1.4.0 version of the dataset has been used for evaluation studies by multiple researchers \cite{ahmedt2019neural, asif2019seizurenet} we also include baseline results of our methods for v1.4.0 to allow a direct performance comparison to these studies. 
In v1.5.2 of the dataset all 7 selected seizure types comprise data from 3 or more patients, which allows statistically meaningful 3-fold patient-wise cross validation. In this scenario, train and test datasets will always contain seizure samples from different patients. This approach makes it more challenging to boost model performance but has higher clinical relevance as it supports model generalisation across patients. For each seizure type, we randomly and equally allocate patients into each fold. We started with seizure types covering less patients and moved on to seizure types carried by more patients. For each seizure type, we exclude patients allocated to previous seizure types. Since datasets of individual patients comprise a different number of seizures, each fold's seizure number can vary largely. Hence we also investigated the impact of selecting different random seeds on the total number of seizures per fold and found that this had essentially no effect on the seizure number for each fold which varied only by plus-minus 3 seizures.

To the best of our knowledge, this is the first seizure type classification study that provides a performance baseline for patient-wise cross validation.


\section{Experiments and Results}
As a first step, to explore the design space in an efficient manner, we chose the two computationally fastest classifiers from Sec. \ref{sec:ml-algos} namely k-NN and SGD classifier and generated their weighted-F1 scores using both pre-processing methods for the first cross validation split. For $f_{max}$, $W_l$, and $O$ i.e. the pre-processing hyperparameters, we generated results for all combinations of $f_{max}$ = \{12, 24, 48, 64, 96\} Hz, $W_l$ = \{1, 2, 4, 8, 16\} secs, and $O$ = \{0.5$W_l$, 0.75$W_l$\} secs. The best hyperparameters of k-NN and SGD for each combination were automatically discovered by running Hyperopt for 100 iterations.  Note that due to the heavy imbalance of the dataset we used weighted-F1 score as scoring metric. 

\begin{figure*}
	\centering
	\includegraphics[width=0.9\textwidth]{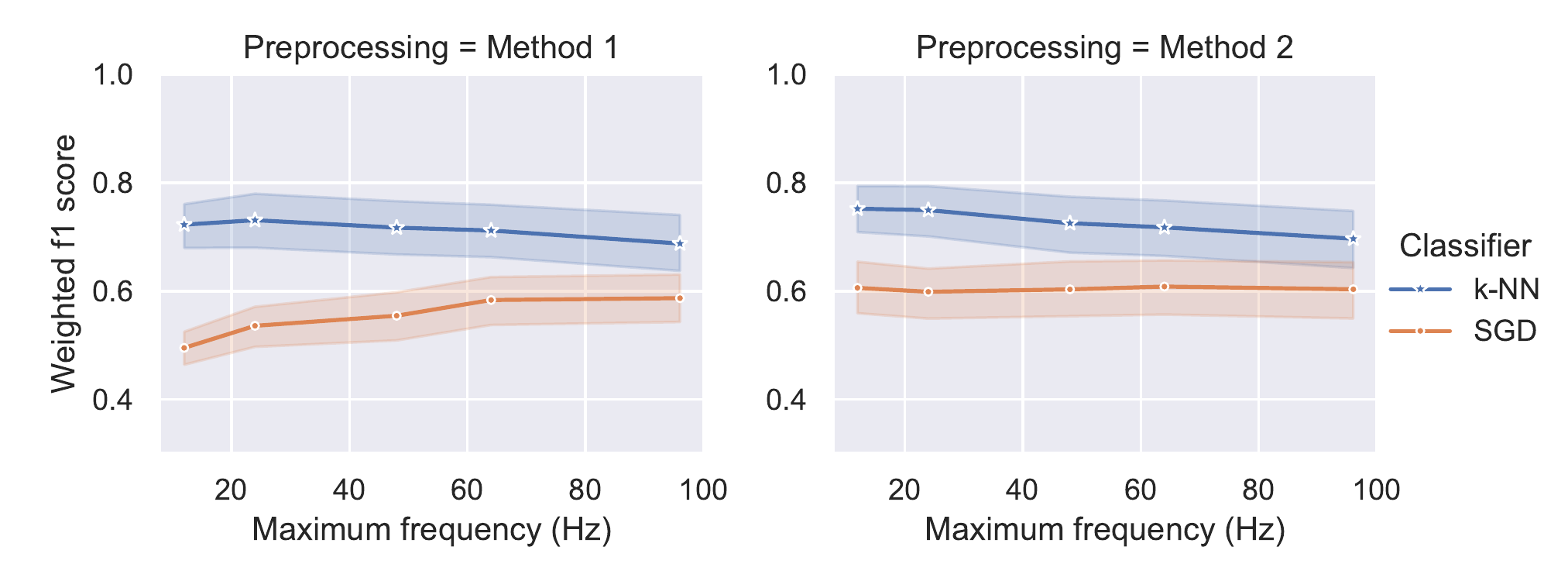} 
	\includegraphics[width=0.9\textwidth]{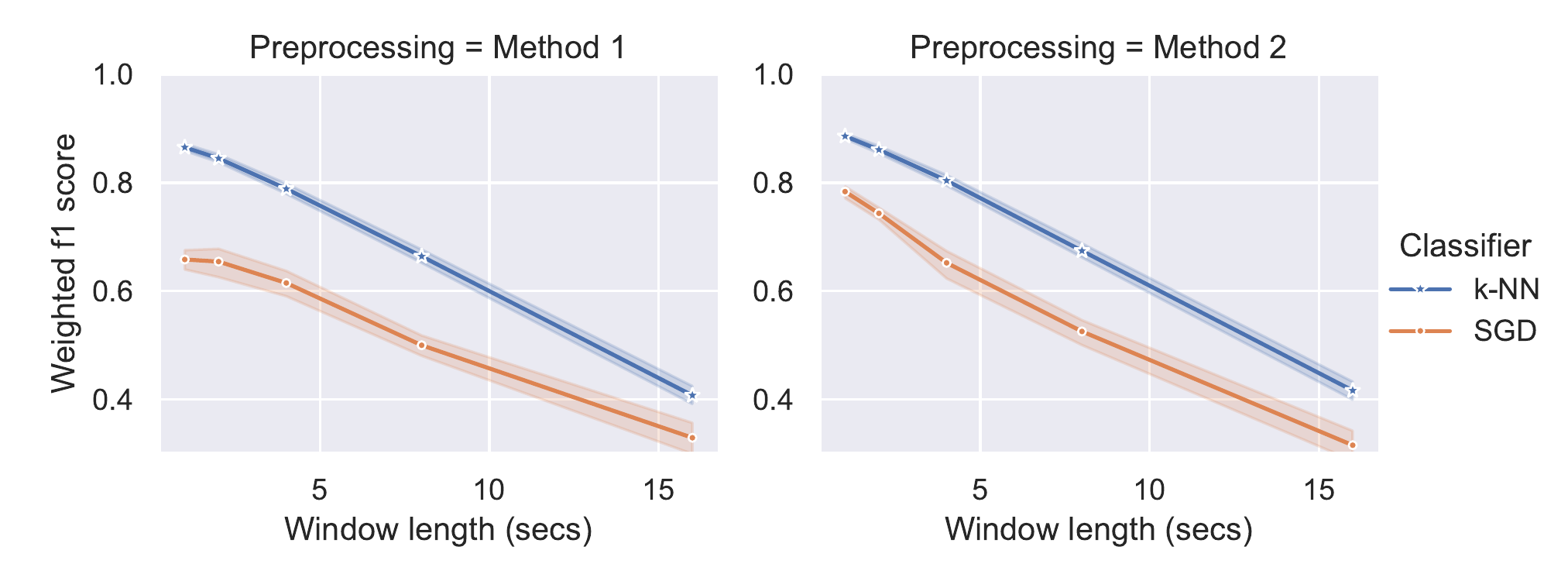}
	\includegraphics[width=0.9\textwidth]{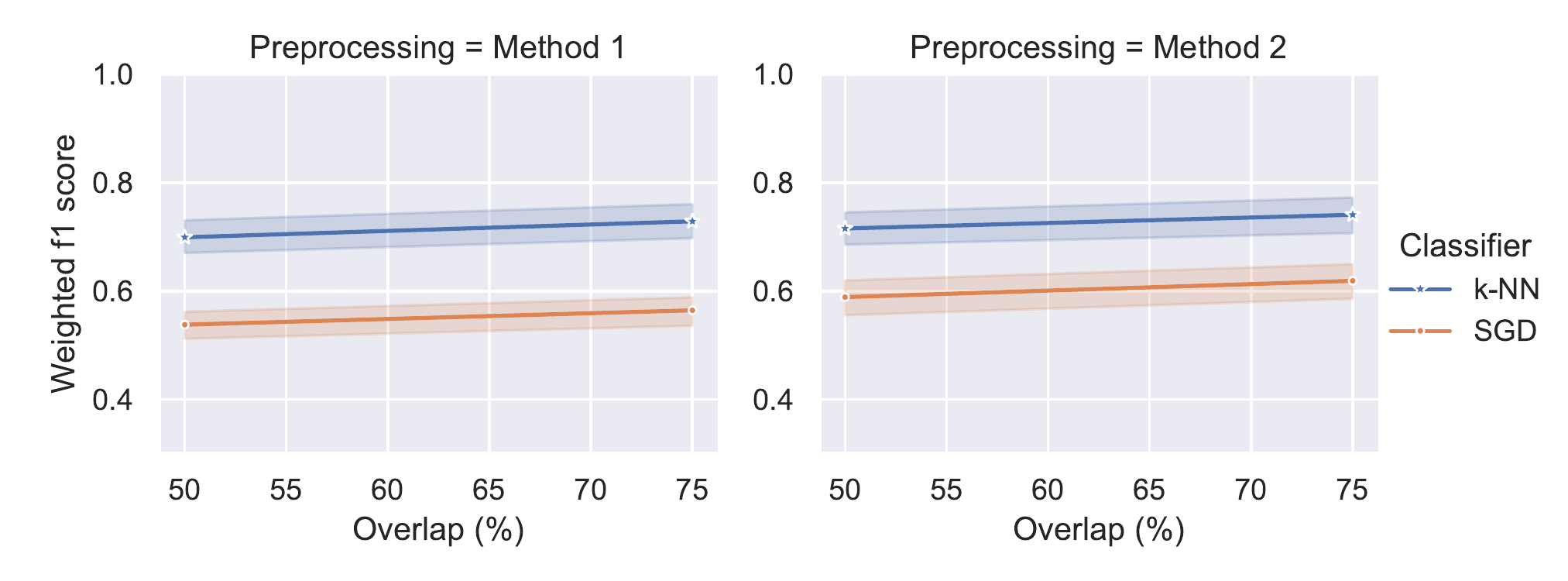}
	\caption{In this figure, we show how the weighted-F1 score varies with $f_{max}$ (top row), $W_l$ (middle row), and $O$ (bottom row) for both pre-processing techniques on k-NN and SGD classifier for v1.4.0} \label{perf_vs_params_v1.4.0}
\end{figure*}

The above experiment served two purposes. Firstly, it allowed us to understand how the performance of the system varies with $f_{max}$, $W_l$ and $O$ separately which is shown in Fig. \ref{perf_vs_params_v1.4.0}. Upon inspecting the top row of Fig. \ref{perf_vs_params_v1.4.0}, we find that while the performance is higher at mid-$f_{max}$ of 24 and 48 Hz, it drops at extreme frequencies. This probably happens since at lower $f_{max}$ we lose relevant information \cite{fisher1992high} and at higher $f_{max}$ the number of dimensions increases, and the classifiers suffer from the curse of dimensionality. The second and third row of Fig. \ref{perf_vs_params_v1.4.0} suggest that the performance increases when $W_l$ decreases and $O$ increases respectively. We speculate that this happens since both the decrease of $W_l$ and increase of $O$ lead to more samples in the training set.

\begin{figure*}
	\centering
	\includegraphics[width=0.9\textwidth]{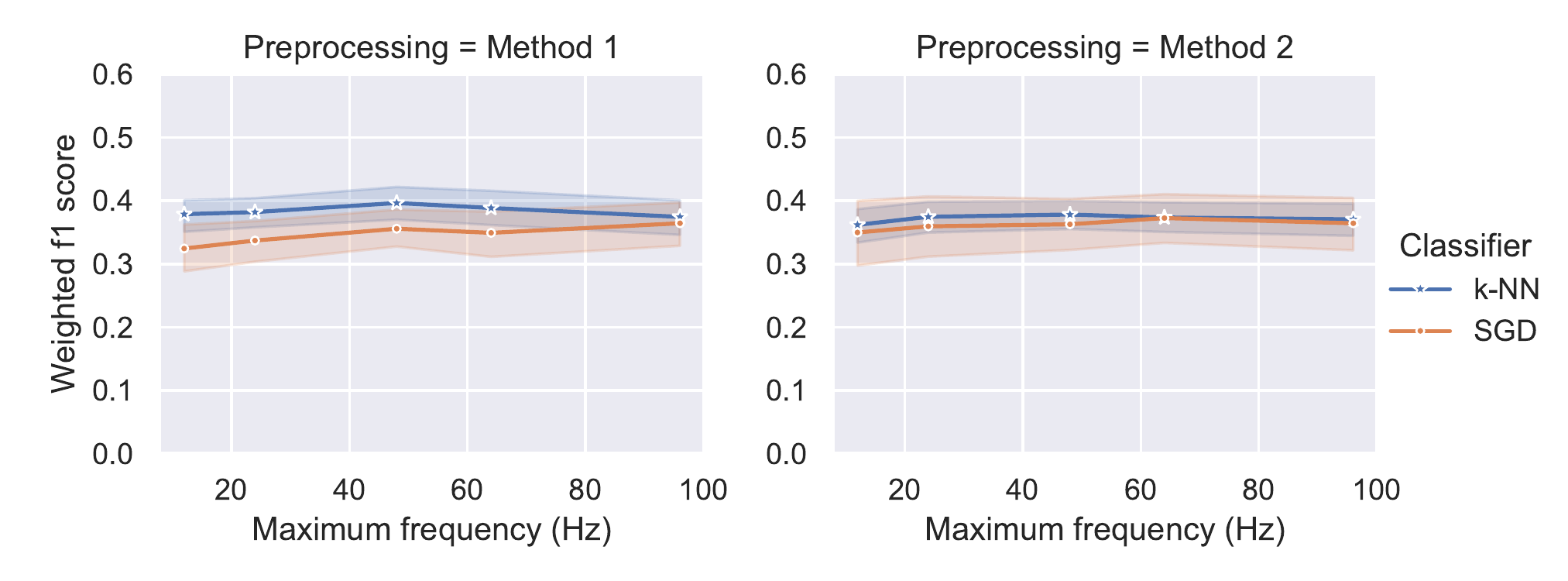}
	\includegraphics[width=0.9\textwidth]{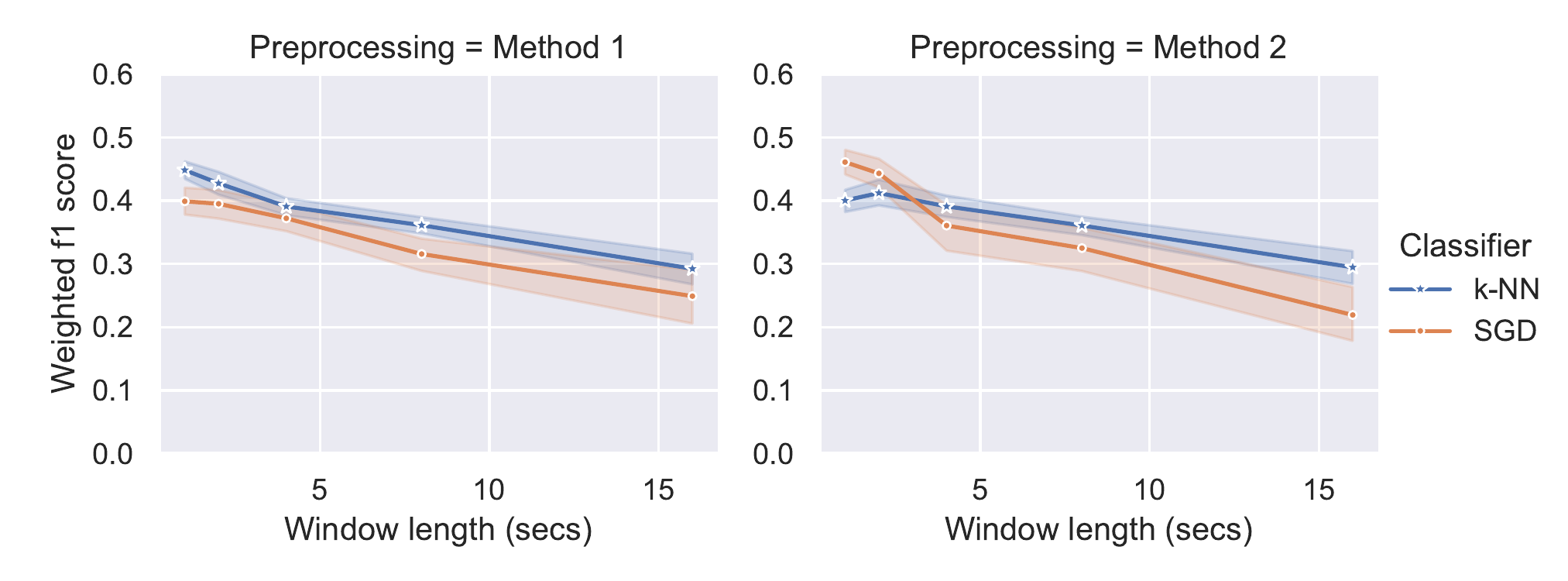}
	\includegraphics[width=0.9\textwidth]{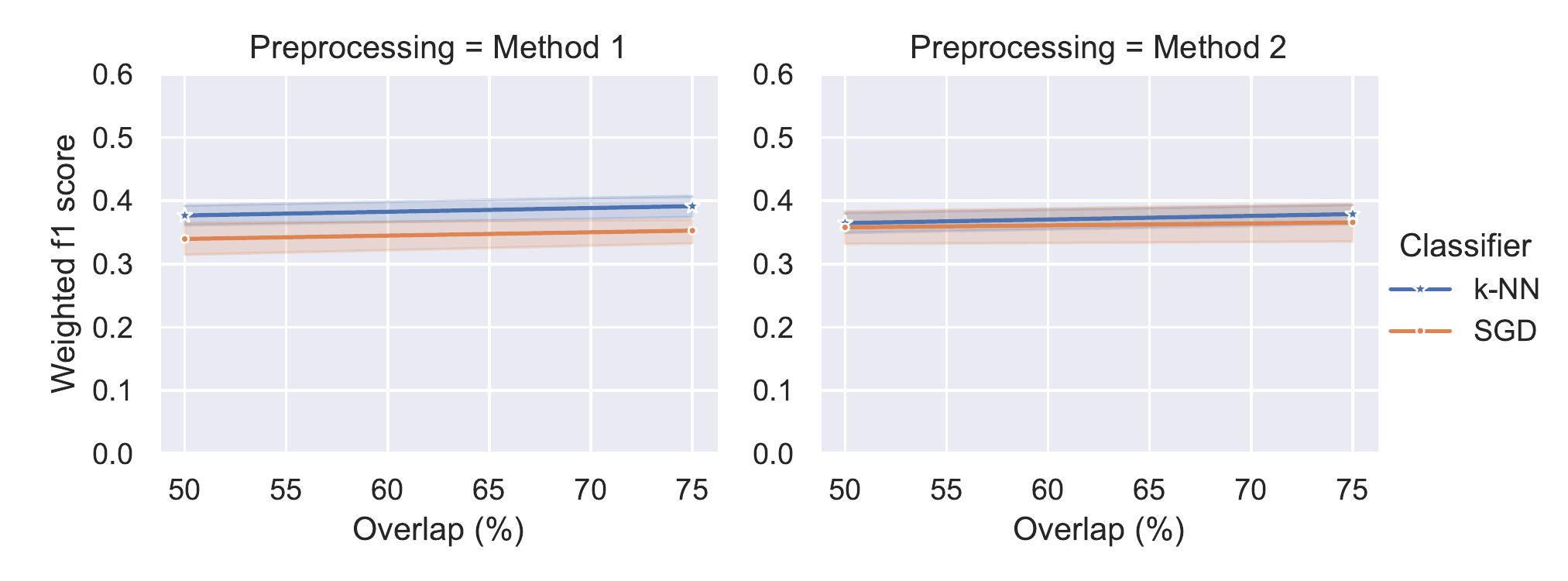}
	\caption{In this figure, we show how the weighted-F1 score varies with $f_{max}$ (top row), $W_l$ (middle row), and $O$ (bottom row) for both pre-processing techniques on k-NN and SGD classifier for v1.5.2} \label{perf_vs_params_v1.5.2}
\end{figure*}

Secondly, this design space exploration using simple classifiers revealed which combination of hyperparameters works best for both pre-processing methods. We select the top four performing sets of hyperparameters and perform 5-fold cross-validation on all the classifiers. Note that CNNs cannot be used to process the data from pre-processing method 2 as it does not produce 2D data. As before, hyperparameters have been chosen by running Hyperopt for 100 iterations. Table \ref{full_results_v1.4.0} shows the obtained average weighted-F1 scores for both pre-processing methods. It is evident that the best performing models were k-NN achieving a weighted-F1 score of $0.901$ for v1.4.0 and XGBoost reaching 0.561 for v1.5.2.

\newcommand{\STAB}[1]{\begin{tabular}{@{}c@{}}#1\end{tabular}}
\begin{table}
	\centering
	\caption{V1.4.0 5-fold seizure-wise cross-validation results on the four top performing hyperparameter sets for each pre-processing method.}
	
	\begin{tabular}{|c|c|c|c|c|c|c|c|}
		\hline
		& \multicolumn{1}{c|}{\GREY $f_{max}$} & \multicolumn{1}{c|}{\GREY $W_l$} & \multicolumn{1}{c|}{\GREY  $O$} & \multicolumn{1}{c|}{\GREY $k-NN$} & \multicolumn{1}{c|}{\GREY $SGD$} & \multicolumn{1}{c|}{\GREY $XGBoost$ } & \multicolumn{1}{c|}{\GREY  $CNN$}\\ \hline
		\multirow{3}{*}{\STAB{\rotatebox[origin=c]{90}{Method 1}}}
		& $48$ & $1$ & $0.75W_l$ & $\textbf{0.884}$ & $0.695$ & $0.817$   & $0.714$ \\ 
		& $24$ & $1$ & $0.75W_l$ & $0.883$ & $0.621$ & $0.844$ &  $0.722$ \\ 
		& $96$ & $1$ & $0.75W_l$ & $0.880$ & $0.724$ & $0.745$ &  $0.718$ \\ 
		& $24$ & $1$ & $0.5W_l$ & $0.879$ & $0.604$ & $0.766$   & $0.713$ \\ 
		\hline
		\multirow{3}{*}{\STAB{\rotatebox[origin=c]{90}{Method 2}}}
        & $48$ & $1$ & $0.75W_l$ & $\textbf{0.901}$ & $0.807$ & $0.851$   & $NA$ \\ 
		& $24$ & $1$ & $0.75W_l$ & $0.900$ & $0.783$ & $0.858$  & $NA$ \\
		& $24$ & $1$ & $0.5W_l$ & $0.895$ & $0.752$ & $0.819$  & $NA$ \\
		& $96$ & $1$ & $0.75W_l$ & $0.890$ & $0.806$ & $0.866$  & $NA$ \\
		\hline
	\end{tabular}\label{full_results_v1.4.0}
\end{table} 

\begin{table}
	\centering
	\caption{V1.5.2 3-fold patient-wise cross-validation results on the four top performing hyperparameter sets for each pre-processing method.}
	
	\begin{tabular}{|c|c|c|c|c|c|c|c|}
		\hline
		& \multicolumn{1}{c|}{\GREY $f_{max}$} & \multicolumn{1}{c|}{\GREY $W_l$} & \multicolumn{1}{c|}{\GREY  $O$} & \multicolumn{1}{c|}{\GREY $k-NN$} & \multicolumn{1}{c|}{\GREY $SGD$} & \multicolumn{1}{c|}{\GREY $XGBoost$ }  &
		\multicolumn{1}{c|}{\GREY  $CNN$}\\ \hline
		\multirow{3}{*}{\STAB{\rotatebox[origin=c]{90}{Method 1}}}
		& $96$ & $1$ & $0.75W_l$ & $0.466$ & $0.432$ & $\textbf{0.561}$ & $0.524$ \\ 
		& $24$ & $1$ & $0.75W_l$ & $0.437$ & $0.384$ & $0.559$ &  $0.530$ \\ 
		& $48$ & $1$ & $0.75W_l$ & $0.467$ & $0.407$ & $0.526$ &$0.525$ \\
		& $24$ & $1$ & $0.5W_l$ & $0.423$ & $0.390$ & $0.512$ & $0.504$ \\ 
		\hline
		\multirow{3}{*}{\STAB{\rotatebox[origin=c]{90}{Method 2}}}
        & $48$ & $1$ & $0.75W_l$ & $0.401$ & $0.469$ & $\textbf{0.542}$ & $NA$ \\
        & $96$ & $1$ & $0.75W_l$ & $0.418$ & $0.459$ & $0.535$ & $NA$ \\
        & $24$ & $1$ & $0.5W_l$ & $0.392$ & $0.452$ & $0.530$ & $NA$ \\
		& $24$ & $1$ & $0.75W_l$ & $0.412$ & $0.462$ & $0.524$ & $NA$ \\
		\hline
	\end{tabular}\label{full_results_v1.5.2}
	\vspace{-1em} 
\end{table} 

The results shown in Table \ref{full_results_v1.4.0} and Table \ref{full_results_v1.5.2} depict that automated seizure type classification is possible using machine learning. $k-NN$ and $XGBoost$ are the best performing algorithms for analysing V1.4.0, and V1.5.2 respectively. We speculate that since V1.5.2 has more seizures compared to V1.4.0 this leads to more training samples and hence paves the path for a more complex algorithm to excel.

Automated detection and classification of seizures is the first step towards building a digital seizure diary which could enable the recording of patient-wise seizure metadata during clinical trials and in epilepsy monitoring units. Such information could then be used to tailor a patient-specific seizure suppression system using optimum medication dosages and suitable medical devices. The methods described in this paper may play an important role for building digital seizure diary technology in the future.

\section{Conclusion}
In this article, we performed the first exploratory study to show that machine learning techniques can be used to classify the type of detected seizures. We hope that automatic classification of seizure types will improve long-term patient care, enabling timely drug adjustments and remote monitoring. To promote research in this topic, we have released our pre-processed datasets for v1.4.0 \cite{ibmtuh2019} and intend to release pre-processed datasets for v1.5.2 and the code we used to generate the presented results.

\newpage
\footnotesize
\bibliographystyle{IEEEbibSPMB}
\bibliography{IEEEabrv,seizure_type}

\begin{thebibliography}{10}
\providecommand{\url}[1]{#1}
\csname url@samestyle\endcsname
\providecommand{\newblock}{\relax}
\providecommand{\bibinfo}[2]{#2}
\providecommand{\BIBentrySTDinterwordspacing}{\spaceskip=0pt\relax}
\providecommand{\BIBentryALTinterwordstretchfactor}{4}
\providecommand{\BIBentryALTinterwordspacing}{\spaceskip=\fontdimen2\font plus
\BIBentryALTinterwordstretchfactor\fontdimen3\font minus
  \fontdimen4\font\relax}
\providecommand{\BIBforeignlanguage}[2]{{%
\expandafter\ifx\csname l@#1\endcsname\relax
\typeout{** WARNING: IEEEtran.bst: No hyphenation pattern has been}%
\typeout{** loaded for the language `#1'. Using the pattern for}%
\typeout{** the default language instead.}%
\else
\language=\csname l@#1\endcsname
\fi
#2}}
\providecommand{\BIBdecl}{\relax}
\BIBdecl

\bibitem{epibasics}
M.~Patty Obsorne Shafer~RN, ``{About Epilepsy: The Basics},''
  \url{https://www.epilepsy.com/learn/about-epilepsy-basics}, 2014.

\bibitem{roy2018deep}
S.~Roy, I.~Kiral-Kornek, and S.~Harrer, ``Deep learning enabled automatic
  abnormal eeg identification,''   \emph{2018 40th Annual International
  Conference of the IEEE Engineering in Medicine and Biology Society
  (EMBC)}.\hskip 1em plus 0.5em minus 0.4em\relax IEEE, 2018, pp. 2756--2759.

\bibitem{kiral2018epileptic}
I.~Kiral-Kornek, S.~Roy, E.~Nurse, B.~Mashford, P.~Karoly, T.~Carroll,
  D.~Payne, S.~Saha, S.~Baldassano, T.~O'Brien \emph{et~al.}, ``Epileptic
  seizure prediction using big data and deep learning: toward a mobile
  system,'' \emph{EBioMedicine}, vol.~27, pp. 103--111, 2018.

\bibitem{harrer2019TIPS}
S.~Harrer, B.~Antony, P.~Shah, and J.~Hu, ``Artificial intelligence for
  clinical trial design,'' \emph{Trends in Pharmacological Sciences}, vol.~40,
  no.~8, pp. 577--591, 2019.

\bibitem{shah2018temple}
V.~Shah, E.~Von~Weltin, S.~Lopez, J.~R. McHugh, L.~Veloso, M.~Golmohammadi,
  I.~Obeid, and J.~Picone, ``The temple university hospital seizure detection
  corpus,'' \emph{Frontiers in neuroinformatics}, vol.~12, p.~83, 2018.

\bibitem{paul2018various}
Y.~Paul, ``Various epileptic seizure detection techniques using biomedical
  signals: a review,'' \emph{Brain informatics}, vol.~5, no.~2, p.~6, 2018.

\bibitem{schindler2007assessing}
K.~Schindler, H.~Leung, C.~E. Elger, and K.~Lehnertz, ``Assessing seizure
  dynamics by analysing the correlation structure of multichannel intracranial
  eeg,'' \emph{Brain}, vol. 130, no.~1, pp. 65--77, 2007.

\bibitem{bergstra2015hyperopt}
J.~Bergstra, B.~Komer, C.~Eliasmith, D.~Yamins, and D.~D. Cox, ``Hyperopt: a
  python library for model selection and hyperparameter optimization,''
  \emph{Computational Science \& Discovery}, vol.~8, no.~1, p. 014008, 2015.

\bibitem{he2016deep}
K.~He, X.~Zhang, S.~Ren, and J.~Sun, ``Deep residual learning for image
  recognition,''   \emph{Proceedings of the IEEE conference on computer vision
  and pattern recognition}, 2016, pp. 770--778.

\bibitem{ahmedt2019neural}
D.~Ahmedt-Aristizabal, T.~Fernando, S.~Denman, L.~Petersson, M.~J. Aburn, and
  C.~Fookes, ``Neural memory networks for seizure type classification,''
  \emph{arXiv}, pp. arXiv--1912, 2019.

\bibitem{asif2019seizurenet}
U.~Asif, S.~Roy, J.~Tang, and S.~Harrer, ``Seizurenet: Multi-spectral deep
  feature learning for seizure type classification,'' \emph{arXiv preprint
  arXiv:1903.03232}, 2019.

\bibitem{fisher1992high}
R.~S. Fisher, W.~Webber, R.~P. Lesser, S.~Arroyo, and S.~Uematsu,
  ``High-frequency eeg activity at the start of seizures.'' \emph{Journal of
  clinical neurophysiology: official publication of the American
  Electroencephalographic Society}, vol.~9, no.~3, pp. 441--448, 1992.

\bibitem{ibmtuh2019}
S.~Roy, ``{IBM Features For Seizure Detection},''
  \url{https://www.isip.piconepress.com/projects/tuh_eeg/downloads/ibm_seizure_preprocessed/},
  2019.

\end{thebibliography}

\end{document}